\documentclass[conference]{IEEEtran}
\IEEEoverridecommandlockouts
\usepackage{amsmath,amssymb,amsfonts}
\usepackage{algorithmic}
\usepackage{graphicx}
\usepackage{textcomp}
\usepackage[dvipsnames]{xcolor}
\usepackage{algorithmic}
\usepackage{algorithm}
\usepackage{subcaption}
\usepackage{dashrule}
\usepackage{tikz,siunitx}
\usepackage{float}
\usepackage{pst-node}
\usepackage{pifont}
\usepackage{array}
\usepackage{url}
\usepackage{fdsymbol}

\usepackage[export]{adjustbox}
\usepackage[utf8]{inputenc}
\definecolor{indianred}{RGB}{205,92,92}
\definecolor{palevioletred}{RGB}{219,112,147}
\definecolor{deeppink}{RGB}{255,20,147}
\definecolor{teal}{RGB}{0,128,128}
\tikzset{cheating dash/.code args={on #1 off #2 distance #3}{
    \pgfmathparse{#3-#1}\let\rest=\pgfmathresult%
    \pgfmathparse{#1+#2}\let\onoff=\pgfmathresult%
    \pgfmathparse{max(floor(\rest/\onoff), 1)}\let\nfullonoff=\pgfmathresult%
    \pgfmathparse{max((\rest-\onoff*\nfullonoff)/\nfullonoff+#2, #2)}\let\offexpand=\pgfmathresult%
    \tikzset{dash pattern=on #1 off {\offexpand pt}}%
}}

\newcommand{\tikzdash}[1]{%
    \tikz[baseline=-.75ex, trim left, trim right=#1]{%
    \draw[line width=1.2pt,cheating dash=on 5pt off 2pt distance #1] (0,0) -- (#1,0);}%
    }

\begin{document}

\title{CAwa-NeRF: Instant Learning of Compression-Aware NeRF Features}

\author{\IEEEauthorblockN{Omnia Mahmoud, Théo Ladune and Matthieu Gendrin}

\IEEEauthorblockA{Orange Innovation, France}

\IEEEauthorblockA{E-mail: omnia.mohamed, theo.ladune, matthieu.gendrin@orange.com}}
 
\maketitle
\thispagestyle{plain}

\begin{abstract}
Modeling 3D scenes by volumetric feature grids is one of the promising directions of neural approximations to improve Neural Radiance Fields (NeRF). Instant-NGP (INGP) introduced multi-resolution hash encoding from a lookup table of trainable feature grids which enabled learning high-quality neural graphics primitives in a matter of seconds. However, this improvement came at the cost of higher storage size. In this paper, we address this challenge by introducing instant learning of compression-aware NeRF features (CAwa-NeRF), that allows exporting the zip compressed feature grids at the end of the model training with a negligible extra time overhead without changing neither the storage architecture nor the parameters used in the original INGP paper. Nonetheless, the proposed method is not limited to INGP but could also be adapted to any model. By means of extensive simulations, our proposed instant learning pipeline can achieve impressive results on different kinds of static scenes such as single object masked background scenes and real-life scenes captured in our studio. In particular, for single object masked background scenes CAwa-NeRF compresses the feature grids down to {6\%} (1.2 MB) of the original size without any loss in the PSNR (33 dB) or down to {2.4\%} (0.53 MB) with a slight virtual loss (32.31 dB).
\end{abstract}
\vspace{0.5cm}
\begin{IEEEkeywords}
Compression, Deep-Learning, Hash-Encoding, Multi-Resolution, Neural Radiance Fields, Rate-Distortion
\end{IEEEkeywords}

\section{Introduction}
Novel view synthesis has recently caught the attention of researchers due to its huge potential use in a wide range of Virtual Reality (VR) and Augmented Reality (AR) applications. It seeks to achieve photo-realistic rendering for a 3D scene at non-captured viewpoints, given a set of images captured from multiple views with assigned camera poses. The exposure of Neural Radiance Fields (NeRF) \cite{nerf} has strongly influenced 3D scenes modeling and novel-views synthesis by the use of learned Deep Neural Networks (DNN) to map each sampled 3D position at a given viewing direction, to its corresponding volume density and view-dependent color. The learning process in NeRF relied on feeding huge number of ray samples to a large-scale Multi-Layer Perceptrons (MLPs) that implicitly represents the scene. Despite the significant achievements of NeRF, it has a massive computational overhead during both training and rendering the scene. Subsequent developments to analyze and improve NeRF \cite{nerf1,nerf2,nerf3,nerf4,nerf5,nerf6,nerf7} have proved their ability to reconstruct numerous 3D scenes using images and corresponding camera poses. However, there is still the burden of the computational complexity and time.

In view of addressing this issue, recent advancements introduced explicit scene representation methods that uses auxiliary voxel grids to geometrically divide the scene into encoded local features, usually referred as Feature Grids. While the actual radiance field representations (density and color) are computed through the use of small-scale MLPs leading to less computational cost. These voxel-based methods have been following various data structures of explicit representations. 

DVGO \cite{dense1} used dense voxel grids to represent the scene, while \cite{dense2} added some CUDA bindings as well as the distortion loss introduced in mip-NeRF \cite{mipnerf} to improve their quality and training speed. In \cite{sparse}, the authors introduced Plenoxels which made use of sparse voxel grids with spherical harmonics to explicitly represent the scene without the use of any MLPs which optimized the training time two orders of magnitudes faster than NeRF. In \cite{tensorf}, they presented TensoRF which is an approach to model radiance fields as a 4D tensor. TensoRF represented the scene as a 3D voxel grid with per voxel multi-channel features. It used a set of vectors and matrices to describe scene appearance and geometry along their corresponding axes which boosted the training time. Moreover, K-Planes \cite{kplanes} further improved the learning quality and speed for a pure PyTorch-based implementations. 

On the other hand, in \cite{sparseoct,plenoctrees} they explicitly represented the scene using 3D octree-based architectures to speed-up the training time over NeRF while preserving the same quality. Nevertheless, the authors in \cite{instant-ngp} came up with a multi-level hash encoding architecture that maps multi-resolution voxel grids into multi-level hash tables of feature grids. The use of hash tables reduced the storage size compared to other voxel grids-based methods and significantly boosted the training time of very high quality models in a matter of seconds. However, there is still the challenge of the required high storage which limits the ability of including NeRFs in some of the applications with storage constrains, particularly streaming, network, and cloud applications.

In this paper, we introduce a compression-aware counter part implementation of INGP, where we focus on learning low entropy features throughout the training. We achieve this by adopting an entropy aware training scheme with a rate-distortion trade-off, inspired by the work done in COOL-CHIC \cite{theo}. To sum up, the paper provides the following contributions: 
\begin{enumerate}
    \item[$\bullet$] The introduction of an entropy-aware 3D features learning technique that could be applied to any NeRF explicit scene representation method regardless of the architecture of the stored features.
     \item[$\bullet$] The per-batch measure of the entropy by directly approximating to a known mathematical probability distribution.
    \item[$\bullet$] The selection of two different continuous probability distributions (Laplace and Cauchy) to measure the global features entropy and evaluate the rate distortion performance for each distribution separately.
    \item[$\bullet$] The rate-loss trade-off adaptation (Adaptive or Hybrid) for performance enhancement according to the type of the chosen probability distribution.
\end{enumerate}
Consequently, CAwa-NeRF achieves outstanding lossless results without sacrificing neither quality nor the super fast training performance of INGP for different types of static scenes compared to the state-of-the-art work listed in Section \ref{sec:sota}. Precisely, for the popular lego scene the feature grids size is scaled down from 21.9 MB to 1.2 MB without loss in the PSNR (even slightly improved from 32.96 dB to 33.14 dB) or down to 0.6 MB with only 0.59 dB loss (32.37 dB). Additionally, with the suitable rate distortion trade-off further lossy compression results could be achieved.

\section{Compressing Neural Fields Related Works}\label{sec:sota}
In light of the huge storage constrains imposed by the accelerated training of explicit voxel grid representations, compression has emerged as a critical point. Several work attempted to reduce the required storage of the learned features. For instance, PlenOctrees \cite{plenoctrees} and Plenoxels \cite{sparse} filtered the voxels storing only a sparse set of the most important voxels, however the size remained notably quite large compared to other models. In \cite{lod}, they learned a multi-resolution tree of feature vectors which can be truncated at any depth to achieve an adaptive bit rate with PSNR penalty. In \cite{compnerf}, the authors learned a hybrid tensor rank decomposition of the scene without neural networks to compress TensoRF \cite{tensorf}. However, their method achieved a near-optimal compression that sacrifices visual loss and training speed. Other work \cite{tinynerf} compressed voxel grids of Plenoxels \cite{sparse} and DVGO \cite{dense1} by applying, frequency domain transformation, pruning, and quantization which reduced the model size by two orders of magnitude with nearly the same PSNR and training speed. 

PlenOctrees \cite{plenoctrees}, PeRFception \cite{perf}, and
Re:NeRF \cite{renerf} all apply bit quantization of the trained local features. Where, Re:NeRF \cite{renerf} and VQRF \cite{1mb} have proposed post-processing compression optimizations pipelines that can achieve significant compression results with the expense of a time consuming overhead limited by the performance of the pre-trained model.

Moreover, other work \cite{varbr,binrf} successfully reduced the size following binary-based features learning. Such that, in \cite{varbr} they proposed vector-quantized auto-decoder (VQ-AD), which adapted soft-indexing on-train quantization to achieve variable bit rates for the feature grids with drawbacks of training footprints of memory and time. On the other hand, in \cite{binrf}, they proposed binary radiance-field (BiRF) which is a binary-based storage-efficient radiance field representations with longer training time compared to INGP with hybrid 2D-3D feature grids explicit representations instead of 3D only.

Finally, SHACIRA \cite{SHACIRA} managed to compress the hash-grid used for implicit representation-based models by the use of an MLP to learn the distribution of the quantized features to minimize its entropy. However, adding more MLPs to the learning model quiet affects the instant learning speed achieved by INGP. 
\section{Prerequisites}
Neural radiance fields (NeRFs) learn an optimized continuous 5D function that maps a 3D point $\mathbf{x} = (x,y,z)$ sampled along the ray $\mathbf{r}(t)$ and the 2D viewing direction $\mathbf{d} = (\theta,\phi)$ to emitted view-dependent color $\mathbf{c} = (r,g,b)$, and volume density $\sigma$ which models the differential probability of ray termination at that point $\mathbf{x}$ \cite{nerf}. The implicit 5D function consists of a large-scale MLP such that,
\begin{equation}
(\mathbf{c},\sigma) = \mathrm{MLP}_\Phi(\mathbf{x},\mathbf{d}),
    \label{eq:nerfmlp}
\end{equation}
where $\Phi$ represents the learnable MLP weights. Afterwards for volumetric rendering, the colors $\mathbf{c}_i$ and the densities $\sigma_i$ of all the sampled points along the ray $\mathbf{r}(t)=\mathbf{o}+t\mathbf{d}$ starting at the ray origin $\mathbf{o}$, are accumulated to obtain the ray color
\begin{equation}
\hat{C}(\mathbf{r})=\sum_{i=1}^N T_i \alpha_i \mathbf{c}_i,
    \label{eq:raycolor}
\end{equation}
where $T_i$ and $\alpha_i$ represent accumulated transmittance and ray termination probability respectively for the $i$-th sampled point and defined as follows,
\begin{equation}
T_i = \prod_{j=1}^i-1 (1-\alpha_j), \alpha_i = 1 - \mathrm{exp}(-\sigma_i \delta_i),
    \label{eq:trans_alpha}
\end{equation}
where $\delta_i$ denotes the distance between adjacent sampled points.

To accelerate the learning, INGP relies on an occupancy grid to skip empty spaces for efficient ray sampling. To further accelerate the learning and rendering processes, a volumetric multi-resolution feature grids representation of the scene was proposed, such that local features were encoded in an explicit data structure of multi-resolution ($L$ levels, ranges from $N_\mathrm{min}$ to $N_\mathrm{max}$ as in \cite{instant-ngp}) hash tables of size $T$ and ${F}$ features per entry. Each hash table $l$ has $\boldsymbol{\theta}_l$ $\in \Theta$ set of features, where $\theta_{l,\tau,f} \in \boldsymbol{\theta_l}$ denotes one feature entry. For each sampled point, the selected features $\mathbf{f} \in \Theta$ mapped by the hash encoded corners points of the sample location inside the multi-resolution voxel grids, are then linearly interpolated per level to compute $\mathbf{y}$ as follows,
\begin{equation}
    \mathbf{y} = \mathrm{interp}(\mathbf{f}), \hspace{0.1cm} \mathrm{where} \hspace{0.3cm} \mathbf{f}= \mathrm{enc}(\mathbf{x};\Theta)
\end{equation}
which is inputted to two small-scale density and colors MLPs $\Phi$ such that,
\begin{equation}
    \sigma = \mathrm{MLP}_{\Phi_1}(\mathbf{y}), \hspace{0.3cm} \mathbf{c} = \mathrm{MLP}_{\Phi_2}(\sigma,\mathbf{d})
    \label{eq:cingp}
\end{equation}

Since the size of the implicit representation MLPs became relatively small, the set of learned explicit hash tables features $\Theta$ significantly affected the learning parameters and constrained the required storage size. Therefore we need to constrain the entropy of the learning parameters referred as feature grids throughout the paper.
\section{Methodology}
\begin{figure*}
    \centering
    \includegraphics[width=0.8\linewidth]{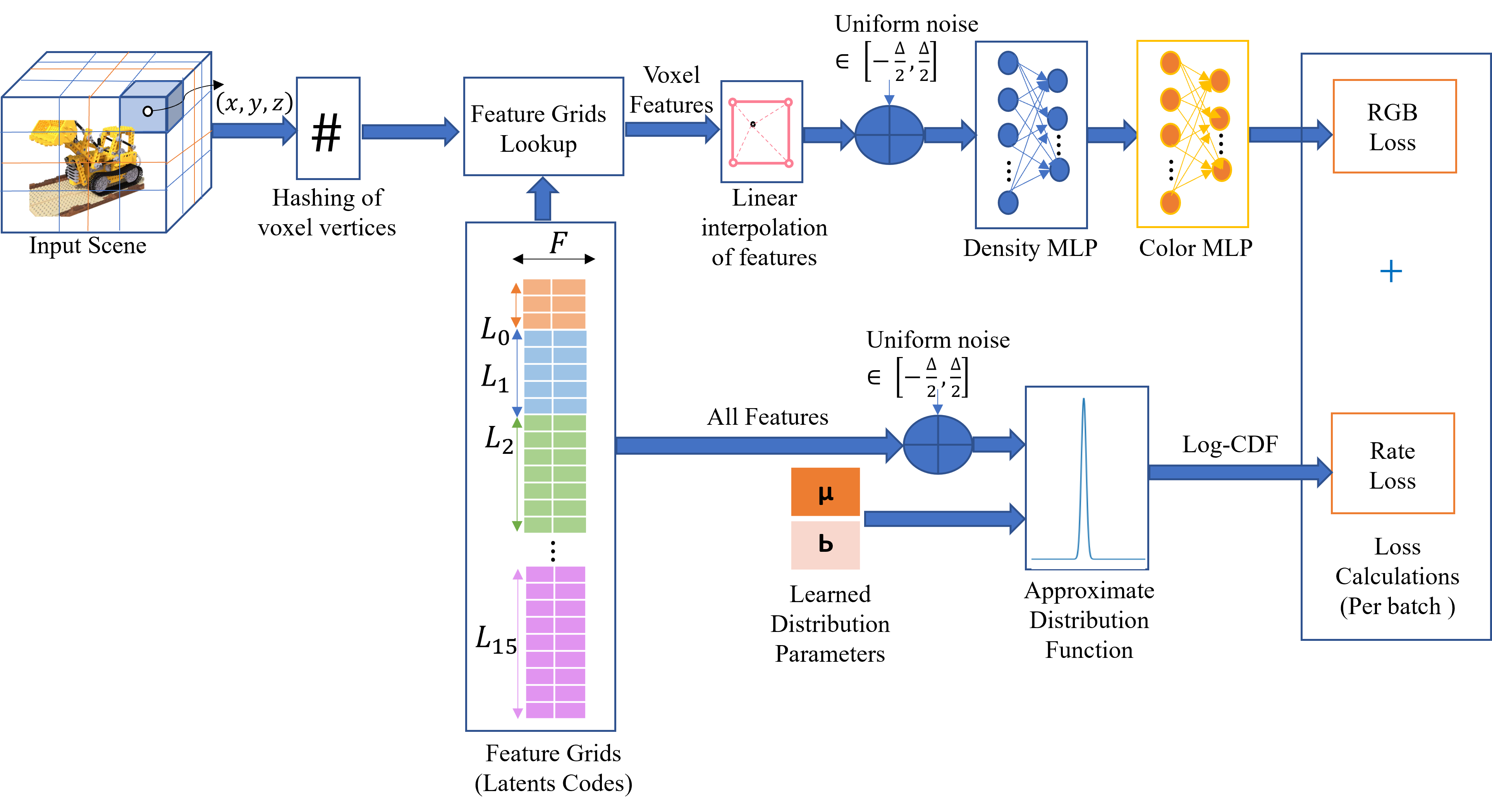}
    \caption{Illustration of the proposed compression-aware features learning method on INGP pipeline.}
    \label{fig:method}
\end{figure*}

This section presents CAwa-NeRF, the proposed method for instant learning of compression-aware feature grids (latents representations) for any explicit scene representation model. Unlike other state-of-the-art compression work, CAwa-NeRF is simple, effective, could be applied to any scene representation model with stored features. However, this paper focuses on the pipeline introduced in INGP \cite{instant-ngp}. Fig \ref{fig:method} shows the overall proposed scheme of learning entropy constrained compression-aware feature grids. We first introduce the proposed compression-aware updates stated in Fig. \ref{fig:method}. Furthermore, we illustrate exporting the zip compressed feature grid from the learned compression-aware feature grids as depicted in Fig. \ref{fig:compr}.

\subsection{Compression-Aware Learning}
To ensure a compression-aware learning of feature grids, the model is prepared for two separate concurrent tasks. The first task is quantization-aware learning and the second task is entropy-aware learning. 
\subsubsection{Quantization-Aware Learning}
The process of quantization itself is not a differentiation friendly operation for gradients calculations during the learning process. However, following the footprints of other work \cite{theo,cNerf}, the noise introduced by quantization could be added during the learning process to the values that needs to be quantized after training. Consequently, we prepare the system for quantization by adding uniform quantization noise on the interpolated features $\mathbf{y}$ such that,
\begin{equation}
    \hat{\mathbf{y}} = \mathbf{y} + \mathbf{n}, \hspace{0.1cm} \mathrm{where} \hspace{0.3cm} \mathbf{n} \sim \mathcal{U}[-\frac{\Delta}{2},\frac{\Delta}{2}],
    \label{eq:yquant}
\end{equation}
where $\Delta$ is the quantization step. Consequently, $\sigma$ in equation (\ref{eq:cingp}) changes to,
\begin{equation}
    \sigma = \mathrm{MLP}_{\Phi_1}(\hat{\mathbf{y}}) 
\end{equation}

It should be noted that adding uniform noise on the input of the MLP does not only immunizing the system against quantization noise, but also against any sort of noise which might comes from either hash table collisions or
the randomized ray batch selection process.
\subsubsection{Entropy-Aware Learning}
To achieve a compression-aware feature grid learning, the approximate feature grids rate term $\mathrm{R}(\Theta)$ is measured. The objective of the learning process becomes minimizing the RGB rendering loss $L_\mathrm{rgb}$ as well as minimizing the amount of information needed to explicitly represent the scene, measured by the rate term $\mathrm{R}(\Theta)$. Therefore, this is modeled by minimizing the following global loss $\mathcal{L}_\mathrm{global}$ per batch of rays ($\mathbf{r} \in \mathcal{R}$),
\begin{equation}
    \mathcal{L}_\mathrm{global} = \mathcal{L}_\mathrm{rgb} + \lambda \mathrm{R}(\Theta),
    \label{eq:lossf}
\end{equation}
where $\lambda$ is the rate distortion loss trade-off that controls the compression to achieve variable bit rate. The RGB loss $L_\mathrm{rgb}$ is calculated as,
\begin{equation}
  \mathcal{L}_\mathrm{rgb} = \sum_{\mathbf{r} \in \mathcal{R}} ||\hat{C}(\mathbf{r})-{C}(\mathbf{r})||_{2}^2,
    \label{eq:rgbloss}
\end{equation}
and the rate loss $\mathrm{R}(\Theta)$ is calculated as follows,
\begin{equation}
    \mathrm{R}(\Theta) = \frac{1}{LTF} \sum_{\theta_{l,\tau,f} \in \Theta} -\mathrm{log}_2 \left[P(\hat{\theta}_{l,\tau,f}+\frac{\Delta}{2})-P(\hat{\theta}_{l,\tau,f}-\frac{\Delta}{2})\right]
\end{equation} 
Accordingly, to calculate the rate loss term $\mathrm{R}(\Theta)$, we directly approximate the distribution of the feature grids to a probability distribution with the cumulative density function (CDF) of $P(\theta)$. Two parameters $\mu$, and $b$ are learned for the distribution. Afterwards, the rate is approximated by the log probability of the quantized features $\hat{\theta}_{l,\tau,f}$ over one quantization interval. The quantized feature $\hat{\theta}_{l,\tau,f}$ is calculated from the learned feature $\theta_{l,\tau,f}$ Similar to equation (\ref{eq:yquant}).

To validate our simple approach of rate measurement, two different continuous probability distributions with a CDF $P(\theta)$ are examined. The first distribution is the Laplace distribution, which has a defined variance and is known by its compressing ability among compression related work \cite{theo}. The two learned variables of the Laplace distribution are $\mu$ and $b$ which represent the mean and the standard-deviation of the distribution respectively.

On the other hand, the other probability distribution is the Cauchy distribution. Unlike the Laplace distribution, the Cauchy distribution does not have a defined variance. Instead, the two parameters $\mu$, and $b$ represent the median and the median absolute deviation (MAD) respectively. Despite that the Laplace distribution has higher tendency to reduce the entropy, the Cauchy distribution has higher weights on the distribution tails which benefits the learned model quality. 

For either distribution, the parameters $\mu$ and $b$ are initialized at the beginning of the training with $\mu_\mathrm{init}$ and $b_\mathrm{init}$ respectively then learned throughout the training.

\subsection{Exporting the compressed Feature-Grid}
\begin{figure*}
    \centering
    \includegraphics[width=0.6\linewidth]{ 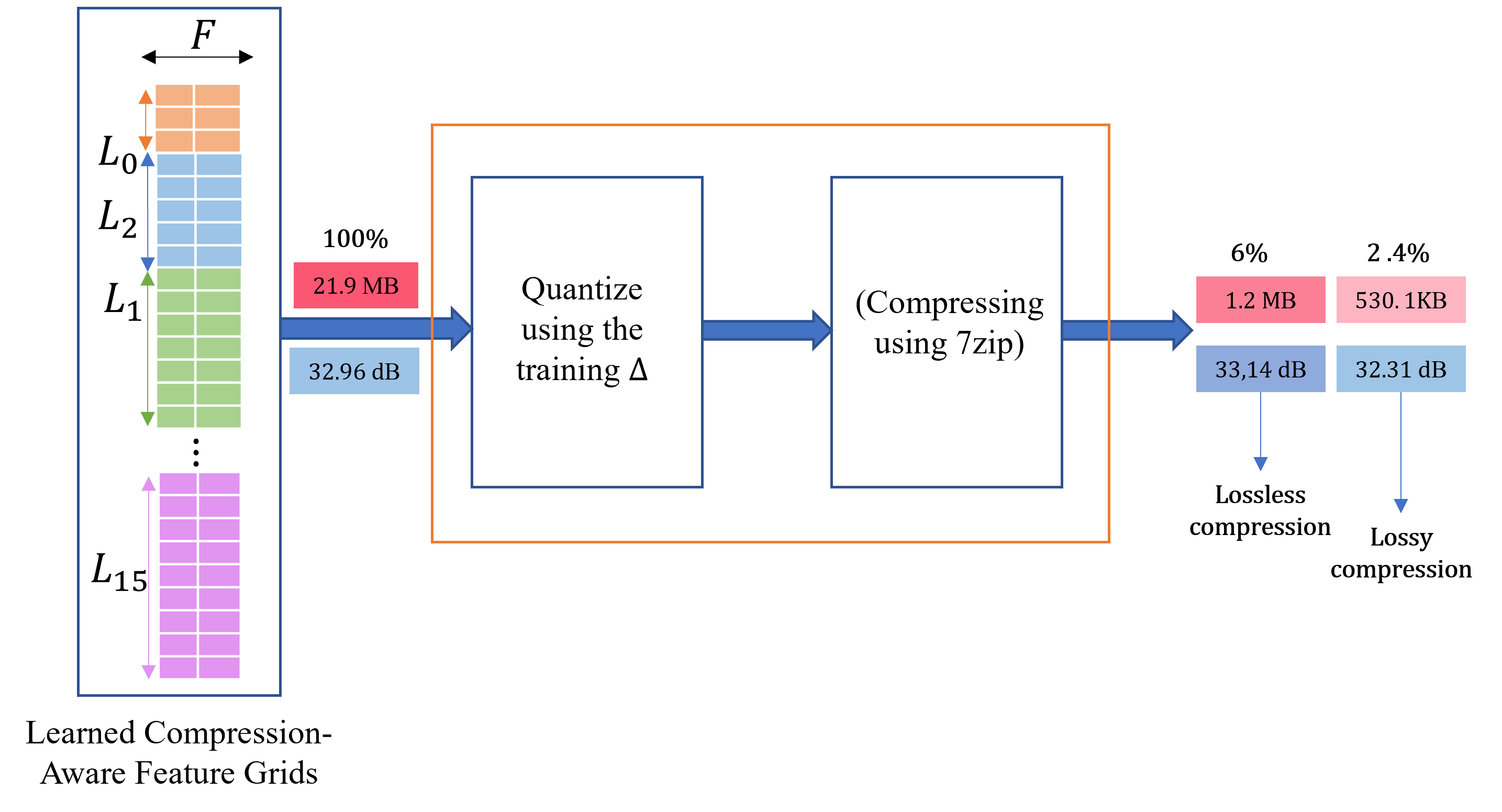}
    \caption{Exporting the compressed features from the learned compression-aware features.}
    \label{fig:compr}
\end{figure*}

After the training stage, the feature grid values are thus learned with minimized entropy. Yet, they are stored in fixed floating point representation which does not provide the maximum benefit from the entropy-aware learning. Therefore, in the following we explain how we export the compressed version of the learned feature grids as depicted in Fig. \ref{fig:compr}.
First we quantize the features according to the type of the distribution used for measuring the rate (whether the distribution is centered or not). Such that, if the Laplace distribution was the one used for rate measure we quantize using the non-centered (mid-rise) quantizer,
\begin{equation}
    \mathrm{Q}_\mathrm{mid-rise}(\theta_{l,\tau,f}) = \Delta \left[ \left\lceil \frac{\theta_{l,\tau,f}}{\Delta} \right\rceil - \frac{1}{2} \right]
\end{equation}
However, if the Cauchy distribution was used for rate measure, we quantize using the following centered (mid-tread) quantizer,
\begin{equation}
    \mathrm{Q}_\mathrm{mid-tread}(\theta_{l,\tau,f}) = \Delta  \left\lfloor \frac{\theta_{l,\tau,f}}{\Delta}  + \frac{1}{2} \right\rfloor,
\end{equation}

Afterwards, we apply any lossless compression tool such as 7zip \cite{7zip} on the quantized feature grids.

It should be noted that the two exporting steps takes only a very few seconds while saving the model. Unlike other time consuming post-processing optimizations proposed in the related work \cite{renerf,1mb}, the time consumed to save our compressed feature grids is negligible.
\section{Experimental Results}
\begin{table}
\centering
\caption{Parameter settings used in CAwa-NeRF experiments and results.}
\resizebox{\columnwidth}{!}{%
\begin{tabular}{l l r} \hline
\textbf{Parameter}&\textbf{Symbol}&\textbf{Value}\\ \hline
Number of levels&$L$& $16$ \\ 
Hash table size&$T$& $2^{19}$\\
Number of features per level&$F$&2\\
Coarsest (min) resolution&$N_\mathrm{min}$& $16$\\ 
Finest (max) resolution&$N_\mathrm{max}$& $2048$\\
Number of training iterations&-& $30000$\\
Quantization step&$\Delta$&0.15\\
Initial distribution parameter 1&$\mu_\mathrm{init}$&0\\
Initial distribution parameter 2&$b_\mathrm{init}$&0.01\\
Fixed rate loss trade-off&$\lambda$& $\{ 1\mathrm{e}^{-4},\dots,5\mathrm{e}^{-3} \}$\\ 
Adaptive rate loss trade-off coefficient&$\bar{\lambda}$&$\{ 0.1,\dots,10 \}$\\
Hybrid rate loss trade-off threshold (Lego)&$\Lambda$&0.0009\\
Hybrid rate loss trade-off threshold (Breakdance)&$\Lambda$&0.005\\\hline
\end{tabular}}
\label{sim}
\end{table}

This section demonstrates the conducted experiments and the achieved results of CAwa-NeRF. All of our experiments are conducted using nerfstudio \cite{nerfstudio}, which provides a simple API for a PyTorch-based end-to-end NeRF simulator with tiny-cuda-nn \cite{tinycuda} bindings. Table \ref{sim} summarizes all the simulation parameters used to obtain the results, unless stated otherwise within the text. The common parameters between our model and INGP are set according to INGP paper. We initialize the distribution parameters $\mu$ and $b$ for to match the small variance of the features at the beginning of the training. Since the features are initialized randomly in the range of $[-10^{-4},10^{-4}]$, they initially should have $\mu_\mathrm{init}=0$. With the fact that we add random noise to the features values before the rate loss calculations to model quantization, we initialize $b=0.01$ to match the very narrow features initialization range. For the quantization step $\Delta$, we chose the maximum quantization step that does not affect the PSNR performance when quantizing the INGP trained feature grids (CAwa-NeRF when $\lambda=0$). 

For all of the provided evaluations, we use the following two scenes to demonstrate the effectiveness of our proposed compression method CAwa-NeRF. The two scenes are chosen to generalize the validity of CAwa-NeRF for both of masked background single object scenes and real-life scenes with different levels of details whether   \newline
\textbf{1. Synthetic (Lego):} Consists of 100 images of 800 × 800 pixel² views taken from either the upper hemisphere or entire sphere around the lego object. We put the last three images aside for evaluating the model and train with the rest. \newline
\textbf{2. Real (Breakdance):} Consists of 34 images of 960x540 pixel² captured with 33 cameras installed in our Orange studio (shown in Fig. \ref{fig:setup}) for a break-dancer shown in Fig. \ref{fig:breakdance}. We put the last three images aside for evaluating the trained model and train with the other 31 images.

Initially, CAwa-NeRF Rate distortion performance is evaluated over a wide range of fixed rate loss trade-off $\lambda \in \{-10^{-4},\dots,5\times -10^{-3} \}$ for both of the considered distributions (Laplace and Cauchy). Afterwards, we propose an adaptive/hybrid rate loss trade-off and verify its effectiveness compared to fixed rate loss trade-off. Following that, we analyze the performance of the proposed method CAwa-NeRF for both of the aforementioned scenes. Last but not least, we evaluate the performance of CAwa-NeRF compared to the most relative state-of-the-art work for the synthetic lego scene.
\subsection{Fixed rate loss trade-off}
\begin{figure}
    \centering
    \includegraphics[width=0.8\linewidth]{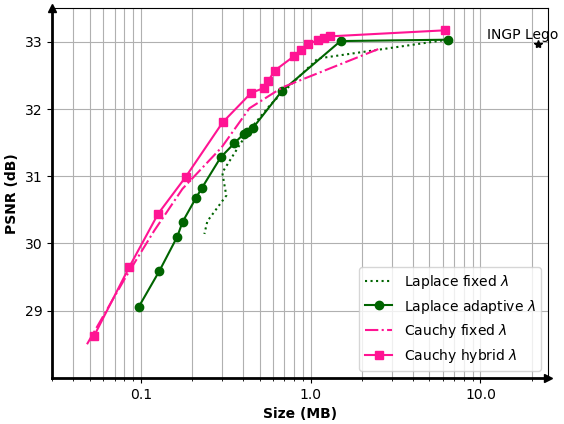}
    \caption{Rate Distortion (PSNR vs. Feature-Grids Size) results of CAwa-NeRF for fixed and updated (adaptive/hybrid) rate loss trade-off $\lambda$ for both of the studied distributions (Laplace, Cauchy).}
    \label{fig:fig1}
\end{figure}

Starting by the Lego scene, we evaluate and compare CAwa-NeRF with both of the Laplace and Cauchy distributions for the range of fixed rate loss trade-off $\lambda$ mentioned in Table \ref{sim} while the other simulation parameters are set as stated in the table. Fig. \ref{fig:fig1} depicts the rate distortion performance for fixed rate loss trade-off for the Laplace and Cauchy distributions in dark green ({\color{OliveGreen}\tikz[baseline]{\draw[line width = 1.5, dotted] (0,.3ex)--++(.3,0);}}) and dark pink ({\color{deeppink}\tikz[baseline]{\draw[line width = 1,dash dot] (0,.4ex)--++(.4,0);}}) respectively. For the Laplace distribution ({\color{OliveGreen}\tikz[baseline]{\draw[line width = 1.5, dotted] (0,.3ex)--++(.3,0);}}), with a minimal rate loss trade-off the feature grid size is scaled down from 6.4 MB to 1.1MB with a slight loss in the PSNR performance. However, to further decrease the size, the PSNR performance aggressively degrades such that the PSNR versus size relation is no more semi-linear.

Unlike Laplace, the Cauchy distribution ({\color{deeppink}\tikz[baseline]{\draw[line width = 1,dash dot] (0,.4ex)--++(.4,0);}}) with the same fixed rate loss trade-off $\lambda$ achieves a remarkable performance that outperforms Laplace with higher rate loss trade-off values (smaller sizes). However, fails to achieve high PSNR qualities for minimal compression similar with small $\lambda$ values. 

Therefore in the following subsection, the rate loss trade-off is updated to an adaptive or hybrid rate loss according to the distribution type to obtain the best rate-distortion results for both of the distributions.

\subsection{Updated adaptive/hybrid rate loss trade-off}
For the Laplace distribution ({\color{OliveGreen}\tikz[baseline]{\draw[line width = 1.5, dotted] (0,.3ex)--++(.3,0);}}), to overcome the aggressive loss on the PSNR especially for higher compression constrains, the rate versus PSNR trade-off $\lambda$ is updated to an adaptive value ($\mathcal{L}_\mathrm{rgb} \bar{\lambda}$). Such that, it changes throughout the training according to the current achieved PSNR quality. Therefore, the loss in equation (\ref{eq:lossf}) becomes,
\begin{equation}
    \mathcal{L}_\mathrm{global}  =  \mathcal{L}_\mathrm{rgb} ( 1 + \bar{\lambda} \mathrm{R}(\Theta)),
    \label{eq:lossfn}
\end{equation}
where, $\bar{\lambda}$ denotes the adaptive rate loss trade-off coefficient. This adaptive rate loss trade-off modifications is done to linearize the relation between the compressed feature grids size and the model quality measured by the PSNR in the semi-log plot, to insure that the entropy of the Laplace distribution is reduced hand-in-hand while improving the model quality. 

On the other hand, the Cauchy distribution naturally has more weight in the distribution tails which favours the PSNR over entropy minimization. Therefor with fixed rate loss ({\color{deeppink}\tikz[baseline]{\draw[line width = 1,dash dot] (0,.4ex)--++(.4,0);}}), it manages to provide substantial results for higher $\lambda$ values (small file sizes), yet has a problem in providing higher PSNR values with small $\lambda$ values (larger file sizes). Therefore, $\lambda$ in equation (\ref{eq:lossf}) is set as in equation (\ref{eq:lossfc}) to be only adaptive at the beginning of the training, to ensure minimal entropy from the beginning of the training and continue prioritizing the PSNR afterwards, yet still provide entropy minimized features.
\begin{equation}
    \lambda = \begin{cases}
              \lfloor \lambda \rceil \hspace{0.1cm} \mathcal{L}_\mathrm{rgb} & \text{if } \lambda < \Lambda < \mathcal{L}_\mathrm{rgb}\\
             \mathcal{L}_\mathrm{rgb}  & \text{if } \Lambda \le \lambda < \mathcal{L}_\mathrm{rgb}\\
             \lambda  & \text{otherwise} \\
       \end{cases},
    \label{eq:lossfc}
\end{equation}
where $\Lambda$ is a threshold for the hybrid rate loss trade-off with the Cauchy distribution only.

To sum up, the Laplace distribution has higher tendency to provide minimal entropy even with the cost of aggressively distorting the quality, therefore an adaptive rate loss trade-off is always needed to maintain the best performance out of the Laplace distribution. While the Cauchy distribution has higher tendency to maintain the best model quality since it has higher weights at the tails of the distribution, therefore an adaptive rate loss trade-off is only needed at the beginning of the training afterwards it should be fixed to provide small sizes with high PSNR at the same time.

Fig. \ref{fig:fig1} depicts the clear improvements of CAwa-NeRF performance for both of, the Laplace distribution with adaptive trade-off $\lambda$ ({\color{OliveGreen}\tikzdash{6pt}\hspace{-2pt}$\medblackcircle$\hspace{-1pt}\tikzdash{6pt}}) with coefficients $\bar{\lambda} \in \{ 0.1,0.5,1,1.1,1.2,\dots,10\}$, as well as the Cauchy distribution with hybrid trade-off $\lambda$ ({\color{deeppink}\tikzdash{6pt}\hspace{-2pt}$\medblacksquare$\hspace{-1pt}\tikzdash{6pt}}). As clearly shown, there is a significant improvement for both of the distribution compared to their fixed $\lambda$. The Laplace distribution with the adaptive rate loss trade-off can reduce the feature grids size from 21.9 MB (INGP) down to 1.3 MB without loss in model quality. Moreover, there is a slight improvement (33.1 dB) over the PSNR performance of the uncompressed INGP model (32.96 dB). Additionally, thanks to the adaptive rate loss trade off in equation (\ref{eq:lossfn}), we can further compress the model without aggressively distorting the PSNR quality. Moreover, the Cauchy distribution outperforms the Laplace distribution, but with the cost of choosing the suitable threshold $\Lambda$. Such that, a compressed features file with size 1.2 MB is achieved with a PSNR improvement to 33.14 dB. On the other hand, for smaller file sizes the Cauchy distribution always outperforms Laplace even with fixed trade-off $\lambda$. It is true that theoretically the Laplace distribution is more compression friendly, however, the higher tendency of the Cauchy distribution to achieved better model qualities eventually serves compression.

\begin{figure}
    \centering
    \includegraphics[width=0.8\linewidth]{ 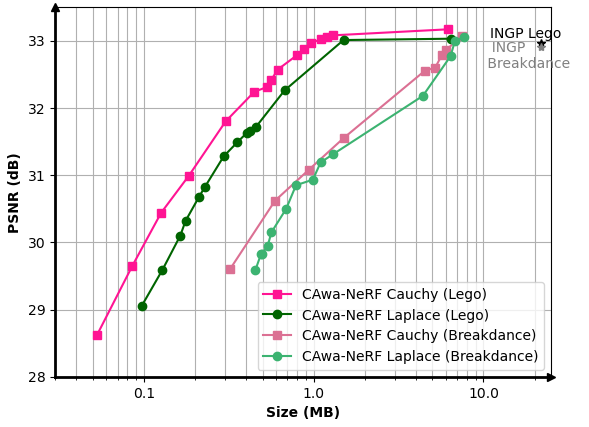}
    \caption{Rate Distortion (PSNR vs. Feature-Grids Size) results of CAwa-NeRF for both of Laplace and Cauchy distributions with their updated rate-loss trade-off $\lambda$ for the synthetic scene (Lego) and the real scene (Breakdance).}
    \label{fig:fig2}
\end{figure}
Now we evaluate CAwa-NeRF performance with the two distributions and their updated trade-offs for captured real life scenes, represented by the breakdance scene. Fig. \ref{fig:fig2} shows the achieved rate distortion results for the Breakdance scene in comparison to the synthetic lego scene, where the uncompressed models for both scenes are plotted in black ({\color{black} $\medstar$}) and gray ({\color{gray} $\medstar$}) for the lego and breakdance scenes respectively. It could be generally noticed that, there is a clear PSNR gap between the compressed lego and breakdance scenes. However this could be explained, since the lego scene has masked background while the breakdance has more information in the background therefore more distortion loss produced by compression. 

In fact for both of the distributions, the real life scene (Breakdance) is compressed from 21.9 MB (INGP) down to 6.4 MB without loss in quality, or down to 1.3 MB with 31.4 dB PSNR for Laplace or down to 1.1 MB with the same PSNR for Cauchy. It is clear that, the Cauchy distribution still outperforms Laplace for the breakdance scenes as well. However, the trade-off threshold $\Lambda$ is set to 0.005. Since the breakdance scene contains more details, the threshold $\Lambda$ needs to be increased compared to the lego in order to relax the rate of entropy minimization and prioritize the PSNR more.

\begin{figure*}
    \centering
    \begin{adjustbox}{minipage=\linewidth,scale=0.7}
    \begin{subfigure}[b]{0.49\linewidth}
    \includegraphics[width=\linewidth]{ 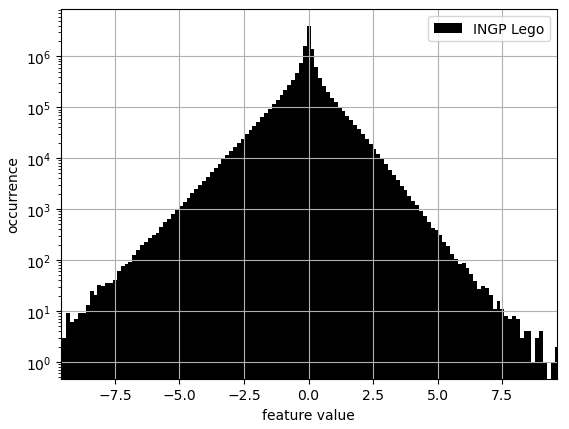}
    \caption{}
    \label{fig:latentsolego}
    \end{subfigure}
    \hfill
    \begin{subfigure}[b]{0.49\linewidth}
    \includegraphics[width=\linewidth]{ 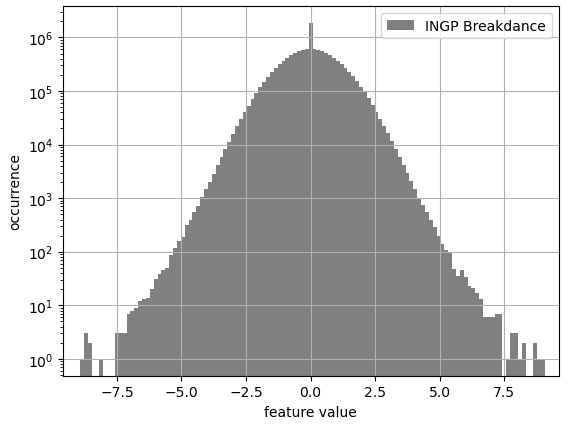}
    \caption{}
    \label{fig:latentsobreakdance}
    \end{subfigure}
    \vfill
     \begin{subfigure}[b]{0.49\linewidth}
    \includegraphics[width=\linewidth]{ 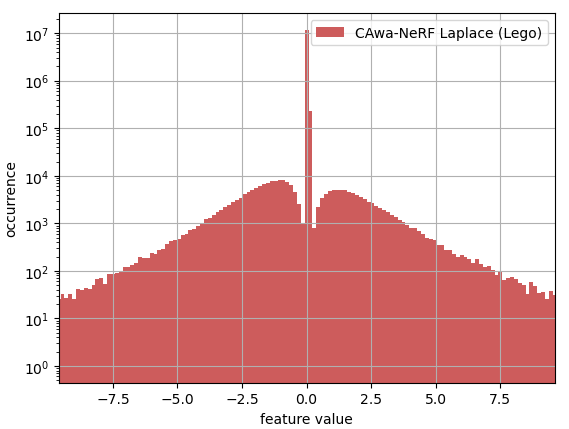}
    \caption{}
    \label{fig:latentsclego}
    \end{subfigure}
    \hfill
    \begin{subfigure}[b]{0.49\linewidth}
    \includegraphics[width=\linewidth]{ 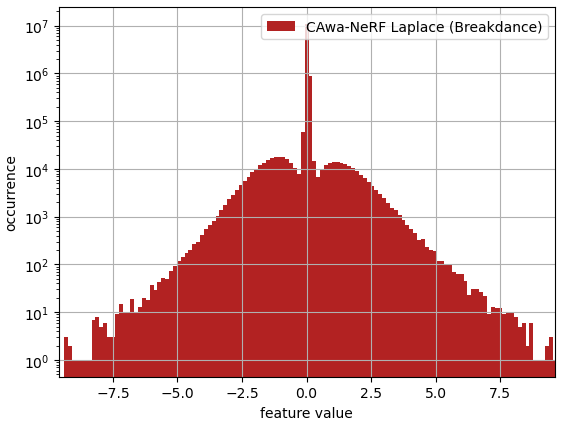}
    \caption{}
    \label{fig:latentscbreakdance}
    \end{subfigure}
    \vfill
     \begin{subfigure}[b]{0.49\linewidth}
    \includegraphics[width=\linewidth]{ 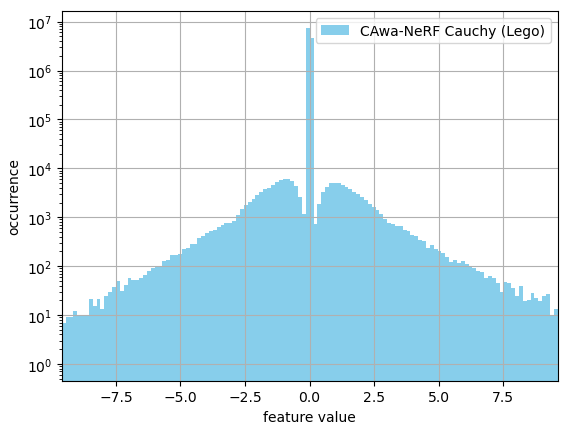}
    \caption{}
    \label{fig:latentsclego_c}
    \end{subfigure}
    \hfill
    \begin{subfigure}[b]{0.49\linewidth}
    \includegraphics[width=\linewidth]{ 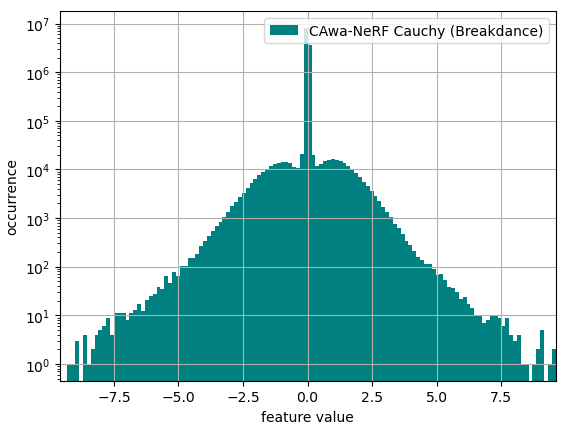}
    \caption{}
    \label{fig:latentscbreakdance_c}
    \end{subfigure}
    \end{adjustbox}
    \caption{Histograms of the learned multi-resolution feature grids: (a) uncompressed INGP Lego. (b) uncompressed INGP Breakdance. (c) CAwa-NeRF Laplace (Lego), $\bar{\lambda}=1$ (PSNR=$31.71$dB, Size=$0.45$MB). (d) CAwa-NeRF Laplace (Breakdance), $\bar{\lambda}=1$ (PSNR=$31.31$dB, Size=$1.1$MB). (e) CAwa-NeRF Cauchy (Lego), ${\lambda}=0.001$ (PSNR=$31.74$dB, Size=$0.31$MB). (f) CAwa-NeRF Cauchy (Breakdance), ${\lambda}=0.0008$ (PSNR=$31.32$dB, Size=$1.1$MB).}
    \label{fig:latents}
\end{figure*}

\begin{figure*}
    \centering
    \begin{adjustbox}{minipage=\linewidth,scale=0.8}
    \begin{subfigure}[b]{0.48\linewidth}
    \includegraphics[width=\linewidth]{ 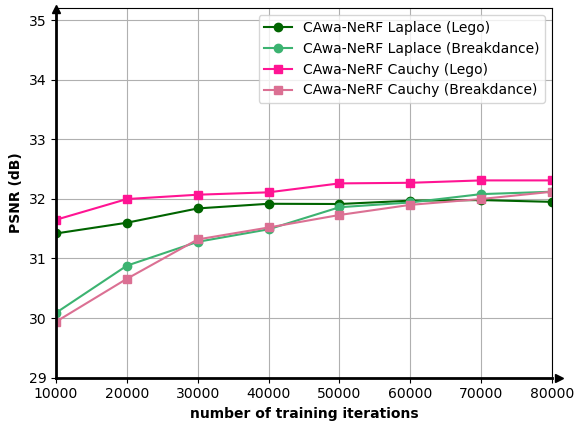}
    \caption{}
    \label{fig:fig4_1}
    \end{subfigure}
    \hfill
    \begin{subfigure}[b]{0.5\linewidth}
    \includegraphics[width=\linewidth]{ 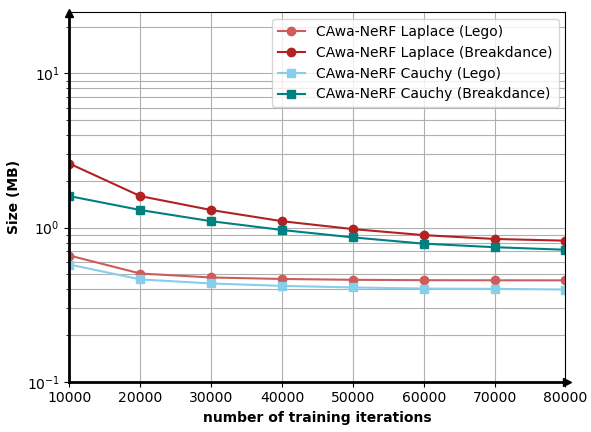}
    \caption{}
    \label{fig:fig4_2}
    \end{subfigure}
    \end{adjustbox}
        \caption{CAwa-NeRF ($\bar{\lambda}=1$) performance for different number of training iterations. (a) The model PSNR plotted in dark and light green for the synthetic scene (lego) and the real scene (breakdance) respectively. (b) Feature grids size plotted in light and dark red for the synthetic scene (lego) and the real scene (breakdance) respectively.}
    \label{fig:fig4}
\end{figure*}

The impact of CAwa-NeRF on the compressed features distributions versus uncompressed (INGP) is visually justified in Fig. \ref{fig:latents}. It depicts a histogram with the occurrence of the features for the quantization step $\Delta=0.15$. We compare the shape of the uncompressed feature grids distributions of both of the synthetic Lego scene and the real Breakdance scene, with their compressed counter parts using both of the Laplace and Cauchy distributions at $\bar{\lambda}=1$ and $\lambda=0.0008$ respectively for almost the same PSNR with either distribution. Generally we can notice a significantly dense distribution for the real scene (breakdance) compared to the synthetic one (lego), which is present since the real scene contains 3D geometric representations of the multiple objects in the whole scene while the synthetic scene has a single object with a masked background. In particular, for the same rate loss trade-off the compressed breakdance scene has almost double the synthetic scene size for the same model quality. 

It is highlighted that, the further compressed the features size, the wider the learned features values range and the narrower the distribution variance around the center unlike the uncompressed case (INGP). Which verifies that throughout the learning process, the actual features distribution is entropy minimized by constraining the entropy directly using either Laplace or Cauchy distributions. It is also depicted that, approximating the rate using the Cauchy distribution yields denser features at the center and less dense at the distribution tails which guarantees smaller sizes over using Laplace distribution for the same PSNR result.

\begin{figure*}
\centering
    \begin{adjustbox}{minipage=\linewidth,scale=0.8}
    \begin{subfigure}[b]{0.48\linewidth}
    \includegraphics[width=\linewidth]{ 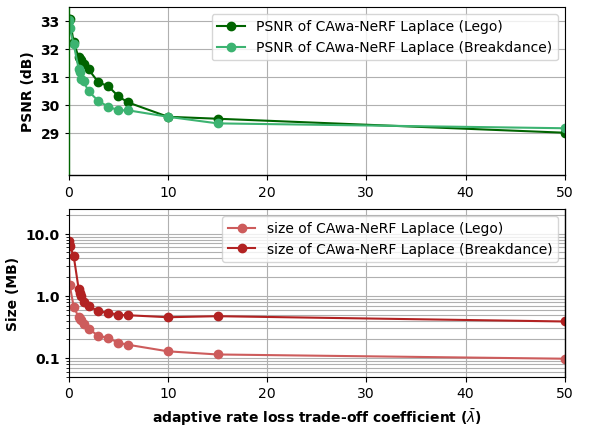}
    \caption{}
    \label{fig:fig3_1}
    \end{subfigure}
    \hfill
    \begin{subfigure}[b]{0.5\linewidth}
    \includegraphics[width=\linewidth]{ 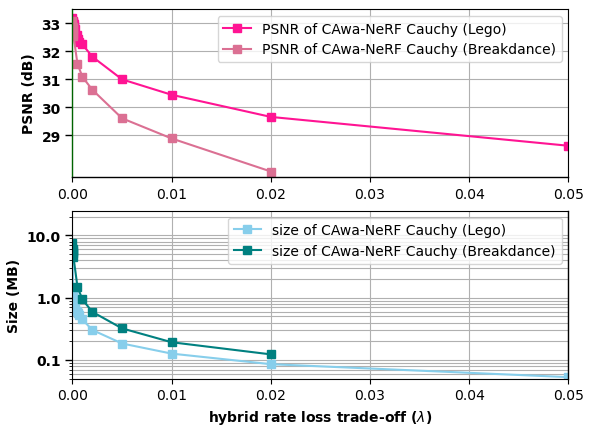}
    \caption{}
    \label{fig:fig3_2}
    \end{subfigure}
    \end{adjustbox}
    \caption{CAwa-NeRF PSNR and compressed features size versus rate loss trade-off ${\lambda}$ for both of the studied scenes and distributions.}
    \label{fig:fig3}
\end{figure*}

To this end, we have been overfitting the model for a fixed number of training iterations as given in Table \ref{sim}. Fig. \ref{fig:fig4} demonstrates the effect of the number of training iterations on the learned model quality (in Fig. \ref{fig:fig4_1}) and the corresponding features size (in Fig. \ref{fig:fig4_2}) for both of the studied distributions and scenes at the same rate loss trade-off. It is generally noticed that, the higher the number of iterations, the lower the file size and the higher the PSNR quality till both saturates. It is also clearly noticeable that the single object scene (lego) saturates faster than the real scene (breakdance). Additionally, the Cauchy distribution always yields the best PSNR and size performance especially for the lego scene. For the real scene, the Laplace distribution converges faster since the Cauchy distribution doesn't have a defined variance so its more sensitive to changes in the PSNR due to compressing the details in the real scene.

To further analyze and highlight the difference between CAwa-NeRF performance for the two studied distributions with their updated rate loss trade-offs, Fig. \ref{fig:fig3_1} depicts the model PSNR ( Lego({\color{OliveGreen}\tikzdash{6pt}\hspace{-2pt}$\medblackcircle$\hspace{-1pt}\tikzdash{6pt}}), Breakdance({\color{ForestGreen}\tikzdash{6pt}\hspace{-2pt}$\medblackcircle$\hspace{-1pt}\tikzdash{6pt}}) ) and the compressed features file size ( Lego({\color{indianred}\tikzdash{6pt}\hspace{-2pt}$\medblackcircle$\hspace{-1pt}\tikzdash{6pt}}), Breakdance({\color{BrickRed}\tikzdash{6pt}\hspace{-2pt}$\medblackcircle$\hspace{-1pt}\tikzdash{6pt}}) ) versus the adaptive rate loss trade-off coefficient $\bar{\lambda}$ for the Laplace distribution. Fig. \ref{fig:fig3_2} shows the PSNR ( Lego({\color{deeppink}\tikzdash{6pt}\hspace{-2pt}$\medblacksquare$\hspace{-1pt}\tikzdash{6pt}}), Breakdance({\color{palevioletred}\tikzdash{6pt}\hspace{-2pt}$\medblacksquare$\hspace{-1pt}\tikzdash{6pt}}) ) and the compressed features file size ( Lego({\color{SkyBlue}\tikzdash{6pt}\hspace{-2pt}$\medblacksquare$\hspace{-1pt}\tikzdash{6pt}}), Breakdance({\color{teal}\tikzdash{6pt}\hspace{-2pt}$\medblacksquare$\hspace{-1pt}\tikzdash{6pt}}) ) versus the hybrid rate loss trade-off $\lambda$ with the Cauchy distribution.

It is generally noticeable that, the higher the rate loss trade-off, the exponentially decaying the size as well as the PSNR. Moreover, the exponential decaying rate of the compressed feature grids size is larger than the PSNR decaying rate such that, for small $\bar{\lambda}$ values there is a noticeable reduction in the size without a severe loss in the PSNR. This explains that the uncompressed model (INGP) considered that all of the features have an equal share in the rendering quality while they do not. For the lego scene, we can notice that the Cauchy distribution provides a less decaying PSNR curve over Laplace. For the breakdance scene, the Cauchy distribution does not significantly improve the PSNR decaying rate, and yet it is still required to select the best threshold $\Lambda$ to achieve the slightly better performance over the Laplace distribution. 

\section{Comparison with related work}
In this section we discuss our achieved CAwa-NeRF compression results compared to the related work mentioned in Section \ref{sec:sota}. To sum up, the previous conducted work to compress NeRF models of explicit scene features representations could be divided into three types. The first one focuses on applying some post-processing steps such as but not limited to, voxel pruning, quantization, and fine tuning on the learned features. These models cannot be directly compared to CAwa-NeRF since we focus on learning compressed features instantly during the training. The second type adapts binary-based learning methods. Such as the proposed auto encoder in \cite{varbr}, where each features entry is learned directly in $n$ bits representation by proposing a soft indexing method for quantization (VQ-AD) to overcome the non-differentiable nature of vector quantization. Another binary-based method is (BiRF) introduced in \cite{binrf} where they binarize each feature entry for the hash table and they add extra hash table for 2D features with size $T_\mathrm{2D}$ as well as the standard 3D features hash table with size $T_\mathrm{3D}$. Finally, entropy minimization based methods which includes SHACIRA \cite{SHACIRA}, where the authors relied into integer quantization as well as minimizing the features entropy learned using an MLP. Table \ref{comp} summarizes our rate distortion achieved results for the synthetic scene compared to both of VQ-AD and BiRF. It is clearly noticeable that CAwa-NeRF is a strong competitor against both of VQ-AD and BiRF especially that it could be integrated to any model not only the hash table based INGP one without a significant increase in the training time. Finally, it still outperforms SHACIRA especially with the ability of reducing the size without any loss in the model PSNR with a much simpler and effective method of entropy constraining. It is worth mentioning that CAwa-NeRF takes only around twice the time needed for training the uncompressed (INGP) model for both of the studied distributions (19 minutes instead of 10 minutes, for 30K iterations).

\begin{table}
\centering
\caption{Comparison between compression related work and CAwa-NeRF in terms of the storage needed for the feature representations versus PSNR for the synthetic NeRF scene (lego). INGP is used as a base-line for the results.}
\resizebox{\columnwidth}{!}{%
\begin{tabular}{c c c} \hline
\textbf{Model Name}&\textbf{PSNR (dB)}&\textbf{Size (MB)}\\ \hline
\rule{0pt}{0.9\normalbaselineskip}INGP&$32.96$&$21.9$\\ \hline
\rule{0pt}{1.05\normalbaselineskip}VQ-AD (6 bits) \cite{varbr}&$30.76$& $0.49$\\
VQ-AD (4 bits) \cite{varbr}&$30.09$&$0.33$\\ \hline
\rule{0pt}{1.05\normalbaselineskip}BiRF ($T_\mathrm{3D}=2^{21}$) \cite{binrf} &$31.61$&$1.94$\\
BiRF ($T_\mathrm{3D}=2^{19}$) \cite{binrf}&$31.53$&$0.72$\\ 
BiRF ($T_\mathrm{3D}=2^{17}$) \cite{binrf} &$30.66$&$0.27$ \\ \hline 
\rule{0pt}{1.05\normalbaselineskip}SHACIRA \cite{SHACIRA} &$32.21$&$1.39$\\ \hline
\rule{0pt}{1.05\normalbaselineskip}CAwa-NeRF Laplace ($\bar{\lambda}=0.2$) &$32.95$&$1.3$\\
CAwa-NeRF Laplace ($\bar{\lambda}=0.5$)&$32.37$&$0.63$\\
CAwa-NeRF Laplace ($\bar{\lambda}=1$)&$31.71$&$0.45$\\
CAwa-NeRF Laplace ($\bar{\lambda}=4$)&$30.82$&$0.22$\\\hline
\rule{0pt}{1.05\normalbaselineskip}CAwa-NeRF Cauchy (${\lambda}=0.00008,\Lambda=0.0009$) &$33.24$&$1.3$\\
CAwa-NeRF Cauchy (${\lambda}=0.0007,\Lambda=0.0009$)&$32.31$&$0.53$\\
CAwa-NeRF Cauchy (${\lambda}=0.002,\Lambda=0.0009$)&$31.74$&$0.31$\\
CAwa-NeRF Cauchy (${\lambda}=0.005,\Lambda=0.0009$)&$31.01$&$0.19$\\\hline
\end{tabular}}
\label{comp}
\end{table}
\section{Conclusion and Future Work}
This paper has proposed an instant learning technique for explicit feature representation NeRF models called compression-aware NeRF (CAwa-NeRF). This is achieved by directly approximating the learned features rate to one of the known continuous probability distributions throughout the model training. Consequently, we have been able to learn minimum entropy and maximum quality feature grids simultaneously with a rate distortion trade-off. We verified that having an adaptive/hybrid rate distortion trade-off dependent on, the model quality throughout the training and the type of the chosen probability distribution, allows further compressing the features with minimal loss in the model quality measured by the PSNR. Moreover, CAwa-NeRF have been tested for different types of scenes and achieved significant performance compared to other conducted work. CAwa-NeRF differentiates itself from other related work in that, it could be integrated to any 3D explicit scene representation model without any changes in the architecture. 

Additionally, CAwa-NeRF is not time consuming unlike previous post-processing based or binary based compression related work, such that it converges to the maximum compression results with minimum entropy in just double the time needed for INGP which is still significantly fast compared to previous work. It is also worth mentioning that the achieved compressed sizes are measured with half precision PyTorch numbers in 16-bit representations per feature entry.

Despite the impressive compression results of CAwa-NeRF on the explicit feature grids representations, we further need to work on compressing the implicit MLPs weights even with their relatively small storage footprint compared to the uncompressed feature grids. We should also examine the compression results for different bit representations per feature other than the fixed 16 bits representation in the paper. Moreover, further work to enhance CAwa-NeRF results by examining other probability models than the Laplace distribution is encouraged. Last but not least, examining how to extent CAwa-NeRF to compressing dynamic scene representations is one of our main interests.
\section*{Acknowledgements}
We would like to thank our Volumetric video team members Christophe Daguet, Sylvain Kedvadec, Cedric Chedaleux, and Stephane Denis for, the numerous fruitful discussions for this project, capturing and calibrating our own testing datasets on the studio, and providing technical references needed to facilitate information collection. 
\bibliographystyle{plain}
\bibliography{references.bib}

\begin{thebibliography}{10}

\bibitem{7zip}
{7-Zip} lossless compression file archiver.
\newblock \url{http://7-zip.org/}.
\newblock Accessed: 2023-07-30.

\bibitem{nerfstudio}
nerfstudio: a simple api of a simplified end-to-end process of creating,
  training, and visualizing nerfs.
\newblock \url{http://docs.nerf.studio/en/latest/index.html}.
\newblock Accessed: 2023-07-30.

\bibitem{tinycuda}
Tiny cuda neural networks.
\newblock \url{http://github.com/NVlabs/tiny-cuda-nn/blob/master/README.md}.
\newblock Accessed: 2023-07-30.

\bibitem{nerf2}
Jonathan~T Barron, Ben Mildenhall, Dor Verbin, Pratul~P Srinivasan, and Peter
  Hedman.
\newblock Mip-nerf 360: Unbounded anti-aliased neural radiance fields.
\newblock In {\em Proceedings of the IEEE/CVF Conference on Computer Vision and
  Pattern Recognition}, pages 5470--5479, 2022.

\bibitem{mipnerf}
Jonathan~T Barron, Ben Mildenhall, Dor Verbin, Pratul~P Srinivasan, and Peter
  Hedman.
\newblock Mip-nerf 360: Unbounded anti-aliased neural radiance fields.
\newblock In {\em Proceedings of the IEEE/CVF Conference on Computer Vision and
  Pattern Recognition}, pages 5470--5479, 2022.

\bibitem{cNerf}
Thomas Bird, Johannes Ball{\'e}, Saurabh Singh, and Philip~A Chou.
\newblock 3d scene compression through entropy penalized neural representation
  functions.
\newblock In {\em 2021 Picture Coding Symposium (PCS)}, pages 1--5. IEEE, 2021.

\bibitem{tensorf}
Anpei Chen, Zexiang Xu, Andreas Geiger, Jingyi Yu, and Hao Su.
\newblock Tensorf: Tensorial radiance fields.
\newblock In {\em European Conference on Computer Vision}, pages 333--350.
  Springer, 2022.

\bibitem{renerf}
Chenxi~Lola Deng and Enzo Tartaglione.
\newblock Compressing explicit voxel grid representations: fast nerfs become
  also small.
\newblock In {\em Proceedings of the IEEE/CVF Winter Conference on Applications
  of Computer Vision}, pages 1236--1245, 2023.

\bibitem{kplanes}
Sara Fridovich-Keil, Giacomo Meanti, Frederik~Rahb{\ae}k Warburg, Benjamin
  Recht, and Angjoo Kanazawa.
\newblock K-planes: Explicit radiance fields in space, time, and appearance.
\newblock In {\em Proceedings of the IEEE/CVF Conference on Computer Vision and
  Pattern Recognition}, pages 12479--12488, 2023.

\bibitem{SHACIRA}
Sharath Girish, Abhinav Shrivastava, and Kamal Gupta.
\newblock Shacira: Scalable hash-grid compression for implicit neural
  representations.
\newblock In {\em Proceedings of the IEEE/CVF International Conference on
  Computer Vision}, pages 17513--17524, 2023.

\bibitem{nerf4}
Yuan-Chen Guo, Di~Kang, Linchao Bao, Yu~He, and Song-Hai Zhang.
\newblock Nerfren: Neural radiance fields with reflections.
\newblock In {\em Proceedings of the IEEE/CVF Conference on Computer Vision and
  Pattern Recognition}, pages 18409--18418, 2022.

\bibitem{perf}
Yoonwoo Jeong, Seungjoo Shin, Junha Lee, Chris Choy, Anima Anandkumar, Minsu
  Cho, and Jaesik Park.
\newblock Perfception: Perception using radiance fields.
\newblock {\em Advances in Neural Information Processing Systems},
  35:26105--26121, 2022.

\bibitem{sparse}
Ratish Jha, Sakshi Seth, and Likitha Karnati.
\newblock Plenoxels: Radiance fields without neural networks.
\newblock 2021.

\bibitem{theo}
Th{\'e}o Ladune, Pierrick Philippe, F{\'e}lix Henry, and Gordon Clare.
\newblock Cool-chic: Coordinate-based low complexity hierarchical image codec.
\newblock {\em arXiv preprint arXiv:2212.05458}, 2022.

\bibitem{1mb}
Lingzhi Li, Zhen Shen, Zhongshu Wang, Li~Shen, and Liefeng Bo.
\newblock Compressing volumetric radiance fields to 1 mb.
\newblock In {\em Proceedings of the IEEE/CVF Conference on Computer Vision and
  Pattern Recognition}, pages 4222--4231, 2023.

\bibitem{sparseoct}
Lingjie Liu, Jiatao Gu, Kyaw Zaw~Lin, Tat-Seng Chua, and Christian Theobalt.
\newblock Neural sparse voxel fields.
\newblock {\em Advances in Neural Information Processing Systems},
  33:15651--15663, 2020.

\bibitem{nerf}
Ben Mildenhall, Pratul~P Srinivasan, Matthew Tancik, Jonathan~T Barron, Ravi
  Ramamoorthi, and Ren Ng.
\newblock Nerf: Representing scenes as neural radiance fields for view
  synthesis.
\newblock {\em Communications of the ACM}, 65(1):99--106, 2021.

\bibitem{instant-ngp}
Thomas M\"{u}ller, Alex Evans, Christoph Schied, and Alexander Keller.
\newblock Instant neural graphics primitives with a multiresolution hash
  encoding.
\newblock {\em ACM Trans. Graph.}, 41(4), 7 2022.

\bibitem{nerf7}
Michael Niemeyer, Jonathan~T Barron, Ben Mildenhall, Mehdi~SM Sajjadi, Andreas
  Geiger, and Noha Radwan.
\newblock Regnerf: Regularizing neural radiance fields for view synthesis from
  sparse inputs.
\newblock In {\em Proceedings of the IEEE/CVF Conference on Computer Vision and
  Pattern Recognition}, pages 5480--5490, 2022.

\bibitem{nerf5}
Keunhong Park, Utkarsh Sinha, Jonathan~T Barron, Sofien Bouaziz, Dan~B Goldman,
  Steven~M Seitz, and Ricardo Martin-Brualla.
\newblock Nerfies: Deformable neural radiance fields.
\newblock In {\em Proceedings of the IEEE/CVF International Conference on
  Computer Vision}, pages 5865--5874, 2021.

\bibitem{nerf6}
Keunhong Park, Utkarsh Sinha, Peter Hedman, Jonathan~T Barron, Sofien Bouaziz,
  Dan~B Goldman, Ricardo Martin-Brualla, and Steven~M Seitz.
\newblock Hypernerf: A higher-dimensional representation for topologically
  varying neural radiance fields.
\newblock {\em arXiv preprint arXiv:2106.13228}, 2021.

\bibitem{binrf}
Seungjoo Shin and Jaesik Park.
\newblock Binary radiance fields.
\newblock {\em arXiv preprint arXiv:2306.07581}, 2023.

\bibitem{dense1}
Cheng Sun, Min Sun, and Hwann-Tzong Chen.
\newblock Direct voxel grid optimization: Super-fast convergence for radiance
  fields reconstruction.
\newblock In {\em Proceedings of the IEEE/CVF Conference on Computer Vision and
  Pattern Recognition}, pages 5459--5469, 2022.

\bibitem{dense2}
Cheng Sun, Min Sun, and Hwann-Tzong Chen.
\newblock Improved direct voxel grid optimization for radiance fields
  reconstruction.
\newblock {\em arXiv preprint arXiv:2206.05085}, 2022.

\bibitem{varbr}
Towaki Takikawa, Alex Evans, Jonathan Tremblay, Thomas M{\"u}ller, Morgan
  McGuire, Alec Jacobson, and Sanja Fidler.
\newblock Variable bitrate neural fields.
\newblock In {\em ACM SIGGRAPH 2022 Conference Proceedings}, pages 1--9, 2022.

\bibitem{lod}
Towaki Takikawa, Joey Litalien, Kangxue Yin, Karsten Kreis, Charles Loop, Derek
  Nowrouzezahrai, Alec Jacobson, Morgan McGuire, and Sanja Fidler.
\newblock Neural geometric level of detail: Real-time rendering with implicit
  3d shapes.
\newblock In {\em Proceedings of the IEEE/CVF Conference on Computer Vision and
  Pattern Recognition}, pages 11358--11367, 2021.

\bibitem{compnerf}
Jiaxiang Tang, Xiaokang Chen, Jingbo Wang, and Gang Zeng.
\newblock Compressible-composable nerf via rank-residual decomposition.
\newblock {\em Advances in Neural Information Processing Systems},
  35:14798--14809, 2022.

\bibitem{nerf3}
Dor Verbin, Peter Hedman, Ben Mildenhall, Todd Zickler, Jonathan~T Barron, and
  Pratul~P Srinivasan.
\newblock Ref-nerf: Structured view-dependent appearance for neural radiance
  fields.
\newblock In {\em 2022 IEEE/CVF Conference on Computer Vision and Pattern
  Recognition (CVPR)}, pages 5481--5490. IEEE, 2022.

\bibitem{plenoctrees}
Alex Yu, Ruilong Li, Matthew Tancik, Hao Li, Ren Ng, and Angjoo Kanazawa.
\newblock Plenoctrees for real-time rendering of neural radiance fields.
\newblock In {\em Proceedings of the IEEE/CVF International Conference on
  Computer Vision}, pages 5752--5761, 2021.

\bibitem{nerf1}
Kai Zhang, Gernot Riegler, Noah Snavely, and Vladlen Koltun.
\newblock Nerf++: Analyzing and improving neural radiance fields.
\newblock {\em arXiv preprint arXiv:2010.07492}, 2020.

\bibitem{tinynerf}
Tianli Zhao, Jiayuan Chen, Cong Leng, and Jian Cheng.
\newblock Tinynerf: Towards 100 x compression of voxel radiance fields.
\newblock {\em Proceedings of the AAAI Conference on Artificial Intelligence},
  37(3):3588--3596, 06 2023.

\end{thebibliography}
\begin{figure}[H]
    \centering
    \includegraphics[width=0.8\linewidth]{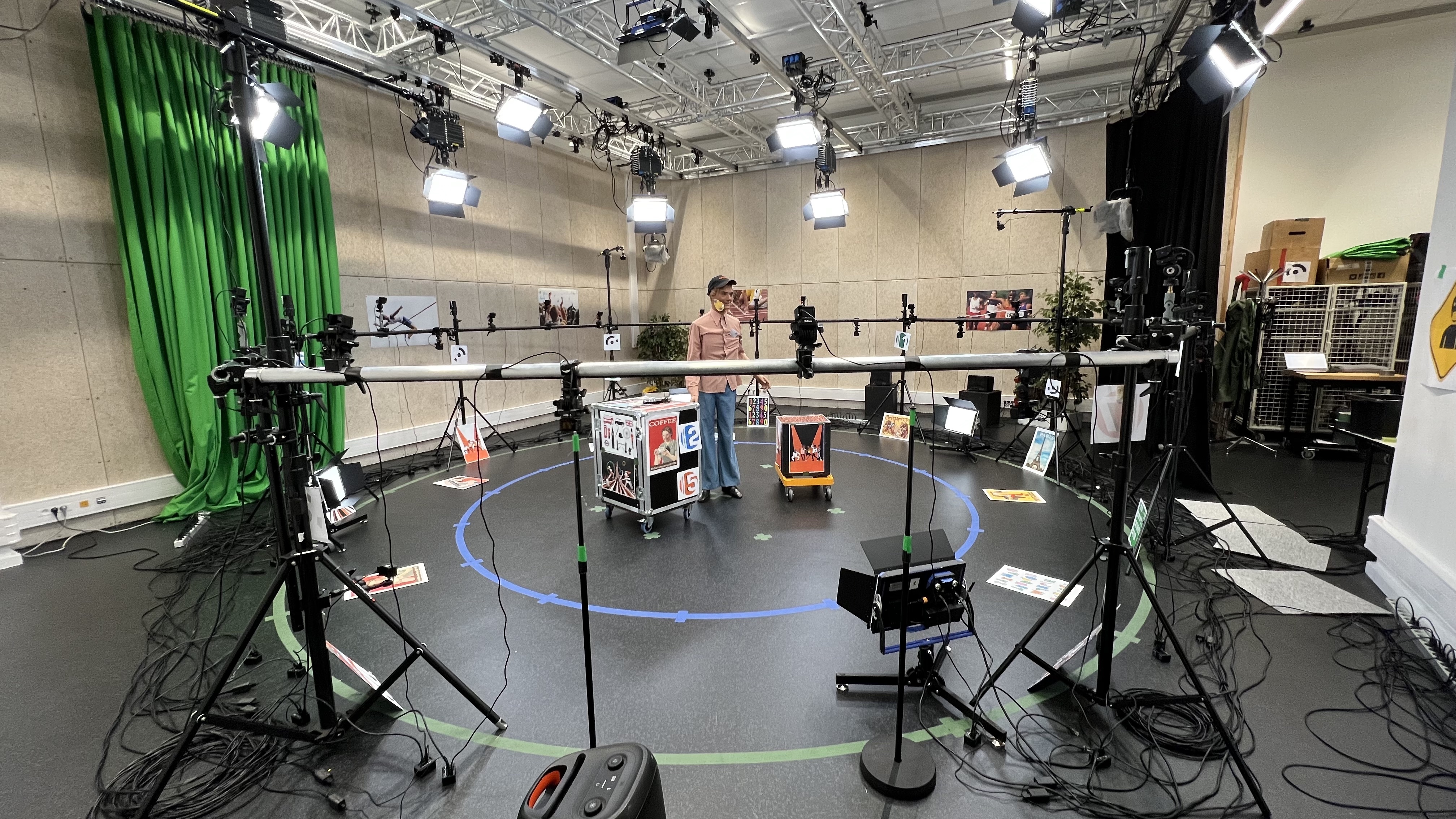}
    \caption{The Camera Setup used to capture the images of the real scene (Breakdance).}
    \label{fig:setup}
\end{figure}
\begin{figure}[H]
    \centering
    \includegraphics[width=0.8\linewidth]{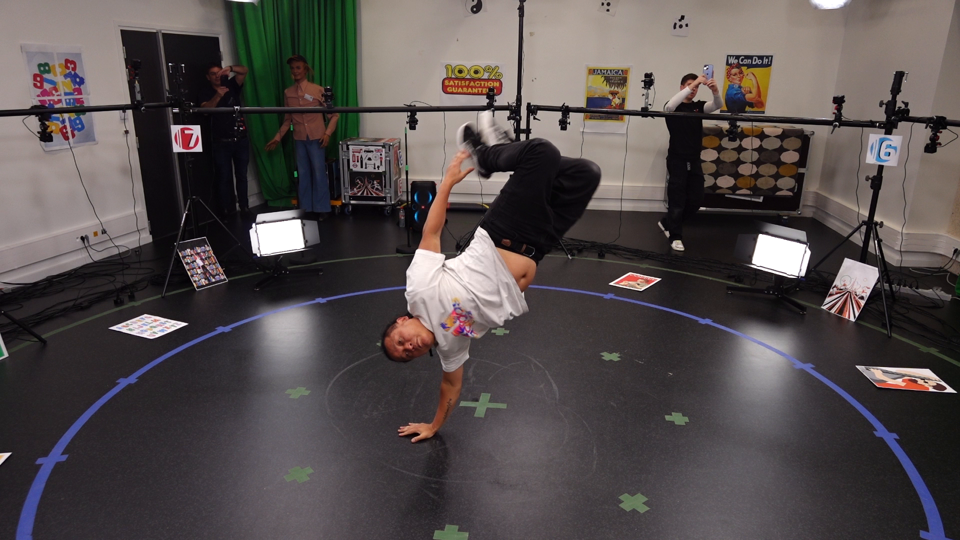}
    \caption{One camera shot for the Breakdance scene used in the conducted experiments.}
    \label{fig:breakdance}
\end{figure}
\end{document}